\documentclass[10pt,twocolumn,letterpaper]{article}

\usepackage[pagenumbers]{wacv} 

\usepackage{graphicx}
\usepackage{amsmath}
\usepackage{amssymb}
\usepackage{booktabs}
\usepackage{stmaryrd}
\usepackage{colortbl}

\usepackage[pagebackref,breaklinks,colorlinks]{hyperref}

\usepackage[capitalize]{cleveref}
\crefname{section}{Sec.}{Secs.}
\Crefname{section}{Section}{Sections}
\Crefname{table}{Table}{Tables}
\crefname{table}{Tab.}{Tabs.}

\def\wacvPaperID{273} 
\def\confName{WACV}
\def\confYear{2025}

\begin{document}

\title{Combining inherent knowledge of vision-language models with unsupervised domain adaptation through strong-weak guidance}

\author{Thomas Westfechtel$^1$, Dexuan Zhang$^1$, Tatsuya Harada$^{1,2}$ \\
$^1$The University of Tokyo \hspace*{1cm} $^2$RIKEN\\
Tokyo, Japan\\
{\tt\small \{thomas,dexuan.zhang,harada\}@mi.t.u-tokyo.ac.jp}}

\maketitle

\begin{abstract}
Unsupervised domain adaptation (UDA) tries to overcome the tedious work of labeling data by leveraging a labeled source dataset and transferring its knowledge to a similar but different target dataset. Meanwhile, current vision-language models exhibit remarkable zero-shot prediction capabilities. 
In this work, we combine knowledge gained through UDA with the inherent knowledge of vision-language models.
We introduce a strong-weak guidance learning scheme that employs zero-shot predictions to help align the source and target dataset. For the strong guidance, we expand the source dataset with the most confident samples of the target dataset. Additionally, we employ a knowledge distillation loss as weak guidance.
The strong guidance uses hard labels but is only applied to the most confident predictions from the target dataset. Conversely, the weak guidance is employed to the whole dataset but uses soft labels. The weak guidance is implemented as a knowledge distillation loss with (adjusted) zero-shot predictions.
We show that our method complements and benefits from prompt adaptation techniques for vision-language models.
We conduct experiments and ablation studies on three benchmarks (OfficeHome, VisDA, and DomainNet), outperforming state-of-the-art methods. Our ablation studies further demonstrate the contributions of different components of our algorithm.
\end{abstract}

\section{Introduction}
\label{sec:intro}

\begin{figure}[t]
   \centering
                \includegraphics[width=0.45\textwidth ]{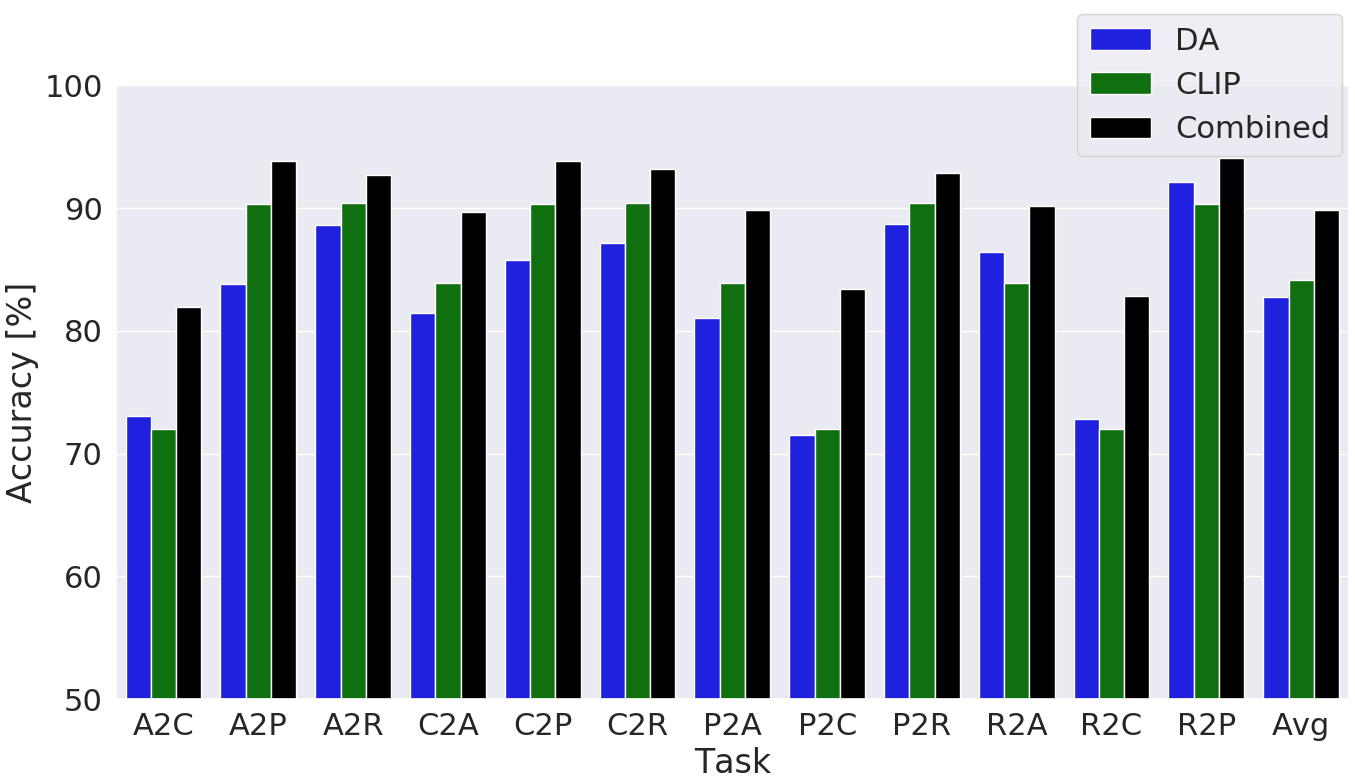}
                \caption{Accuracy on the OfficeHome dataset for unsupervised domain adaptation (blue), CLIP zero-shot predictions (green), and our combined version (black) integrating zero-shot predictions into UDA. In this work, we present a way to combine the knowledge from vision-language models with knowledge transferred via UDA from a source domain. It can be seen that the performance significantly improves.}
                \label{fig:intro}       
  \end{figure}

Deep neural networks have significantly advanced the field of computer vision. However, training these networks requires a large amount of labeled data. Unsupervised domain adaptation (UDA) offers a solution by transferring knowledge from a labeled similar but distinct source dataset to an unlabeled target dataset, reducing the need for extensive labeling.

On the other hand, vision-language models exhibit remarkable zero-shot prediction accuracy, even without any task-specific training data. For instance, on the DomainNet dataset, current vision-language models' zero-shot predictions outperform state-of-the-art domain adaptation without the need for a source dataset. This raises the question of the importance of domain adaption method give the rapid emergence of larger foundation models.

In this work, we argue that rather than seeing these two methods as competing, combining the strengths of both of them achieves even better results. We combine the inherent knowledge of the vision-language models with the knowledge transferred from a source dataset through domain adaptation. 

To preserve the inherent knowledge of the vision-language model we employ a strong-weak guidance. The strong guidance is applied only for the most confident samples using the hard predictions (i.e. pseudo-label) in form of a source dataset expansion. Conversely, the weak guidance employs soft predictions and is applied to all samples in form of a knowledge distillation loss. We show that adjusting the output probabilities by accentuating the winning probabilities enhances the effectiveness of knowledge distillation.

By combining our strong-weak guidance with a conventional unsupervised domain adaptation method we can significantly increase the performance as can be seen in Fig. \ref{fig:intro}. This highlights the potential of combining inherent knowledge of vision-language models with knowledge transferred from a source domain.
	
While much research on domain adaptation with CLIP models has focused on adapting the text prompt, our method focuses on adapting the visual encoder, while maintaining its inherent knowledge. We show that these two approaches are complementary, and show that our method benefits from combining it with the prompt adaptation method DAPL \cite{DAPL}.

Our main contributions are:
\begin{itemize}
\item We introduce a strong-weak guidance scheme to exploit inherent knowledge of vision-language models and combine it with a conventional UDA method.
\item We introduce a combination of weak guidance based on knowledge distillation of adjusted zero-shot predictions and strong guidance based on source dataset expansion. 
\item We show the effectiveness of our algorithm on three datasets (Office-Home, VisDA, and DomainNet) and further evaluate the method in various ablation studies.
\item We show that the method is applicable for both CNN-based and ViT-based backbones and can be combined with other UDA methods.
\end{itemize}

\section{Related works}
\label{sec:formatting}
\subsection{Unsupervised domain adaptation}
The goal of unsupervised domain adaptation is to overcome the domain shift between the source and the target dataset. Surveys for this task can be found in  \cite{s-1}, \cite{s-2}. One approach is to align the feature space distributions between the two datasets. Domain-adversarial neural network (DANN) \cite{DANN} introduces an adversarial approach to achieve this. A domain classifier is trained to distinguish the feature space of source and target samples. The work introduced a gradient reversal layer. This layer inverts the gradients and thereby inverts the training objective, meaning that the backbone is trained to generate features that are indistinguishable for the domain classifier, generating so-called domain invariant features. BIWAA \cite{BIWA} introduces a feature re-weighting approach based on their contribution to the classifier. CDAN \cite{CDAN} extended DANN by multilinear conditioning the domain classifier with the classifier predictions. Moving semantic transfer network \cite{MSTN} introduced a moving semantic loss. 

Another approach aims for the minimization of joint-error \cite{zhang1}. Others approaches employ information maximization or entropy minimization to generate more accurate target predictions. SHOT \cite{SHOT} exploits both information maximization and self-supervised pseudo-labeling to increase the performance of the target domain. SENTRY \cite{SENTRY} uses prediction consistency among different augmentations of an image to either minimize its entropy or maximize it in case of inconsistent predictions.

Following the rise of vision transformers approaches specially tailored for these backbones have gained popularity. TVT \cite{TVT} introduced a Transferability Adaption Module that employs a patch-level domain
discriminator to measure the transferability of patch tokens,
and injects learned transferability into the multi-head self-
attention block of a transformer.
CDTrans \cite{CDTrans} introduced a triple-branch transformer framework with cross-attention between source and target domains.
PMTrans \cite{PMTrans} introduced PatchMix which builds an intermediate domain by mixing patches of source and target images. The mixing process uses a learnable Beta distribution and attention-based scoring function to assign label weights to each patch.

In our work, we employ CDAN \cite{CDAN} as our domain adaptation loss. While the performance is neither state-of-the-art for CNN-based nor transformer-based networks, it achieves reasonable performance for both architectures. \cite{pretrain} has shown that many algorithms developed for CNN backbones underperform for transformer-based backbones. On the other hand, most transformer-based algorithms explicitly make use of the network structure and cannot be employed for CNN-backbones.

\subsection{Vision-language models}
CLIP \cite{CLIP} introduced a contrastive learning strategy between text and image pairs. It employs a text encoder and a visual encoder. Feature representations of positive pairs are pushed together, while unrelated or negative pairs are pushed apart. The usage of visual and language encoders enables zero-shot prediction. The class labels are encoded via the language encoder and serve as classifier. Usually, the class labels are combined with a pretext or ensemble of pretexts such as 'A photo of a \{\textrm{object}\}.' to enhance the performance, where \{\textrm{object}\} is replaced with the respective classes.
While CLIP collected the training dataset by constructing an allowlist of high-frequency visual concepts from Wikipedia, ALIGN \cite{ALIGN} does not employ this step, but makes up for it by using a much larger, though noisier, dataset. 
BASIC \cite{BASIC} further increased the zero-shot prediction capabilities by further increasing data size, model size, and batch size.
LiT \cite{LiT} employs a pre-trained image encoder as visual backbone and aligns a text encoder to it. The vision encoder is frozen during the alignment.
DFN \cite{DFN} investigates data filtering networks to pool high-quality data from noisy curated web data.

In our work, we employ the CLIP pre-trained models due to their popularity. We employ the seven ImageNet templates subset published on the CLIP Git Hub.

 \begin{figure*}[t]
   \centering
                \includegraphics[width=0.6\textwidth ]{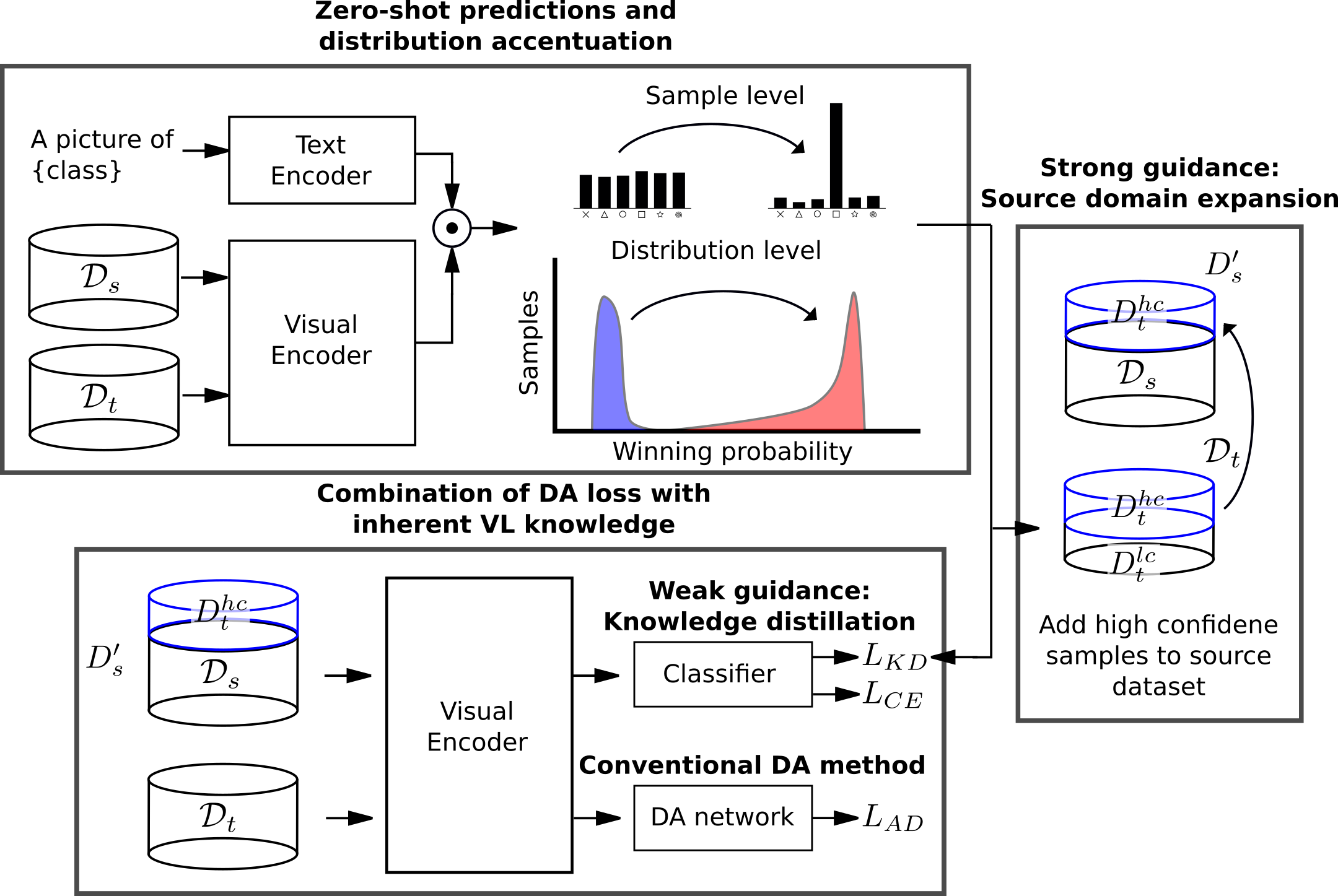}
                \caption{Process flow of our algorithm. In a first step, the zero-shot predictions of source $\mathcal D_s$ and target $\mathcal D_t$ dataset are estimated. We shift the zero-shot predictions distribution through a temperature parameter in the softmax to accentuate the winning probability. Based on the zero-shot predictions, we extend the source dataset with high confident target samples. These samples are treated as source data, with their respective pseudo-labels and represent the strong guidance of our method. The network is then trained using a classification loss $L_{CE}$ for the (expanded) source data, a knowledge distillation loss $L_{KD}$ employing the shifted zero-shot predictions $\tilde y_o$, and an adversarial adaptation loss $L_{DA}$. The knowledge distillation loss represents the weak guidance of the method, as it is employed for all samples and uses the soft zero-shot predictions.}
                \label{fig:flow}       
  \end{figure*}

\subsection{Domain adaptation for vision-language models}
Surprisingly there has not been much research on combining domain adaptation to vision-language models.

Few approaches make use of prompt learning \cite{coop}. This approach trains the prompt’s context words for the text encoder with learnable vectors while the encoder weights are frozen. 
DAPL \cite{DAPL} employs a prompt learning approach to learn domain-agnostic context variables and domain-specific context variables for unsupervised domain adaptation. This approach freezes the text and vision encoder during training.
AD-CLIP \cite{ADCLIP} introduces a prompt learning scheme that entirely leverages the visual encoder of CLIP and introduces a small set of learnable projector networks.
PADCLIP \cite{PADCLIP} aims for a debiasing zero-shot predictions and employs a self-training paradigm, we employ adjusted prediction scores instead of hard pseudo-labels and use an adversarial approach to overcome the domain gap.
EUDA \cite{EMP} employs prompt task-dependant tuning and a visual feature refinement module in combination with pseudo-labeling from semi-supervised classifiers.
MPA \cite{MPA} takes a similar approach of prompt learning for multi-source unsupervised domain adaptation. In a first step, individual prompts are learned to minimize the domain gap through a contrastive loss. Then, MPA employs an auto-encoding process and aligns the learned prompts by maximizing the agreement of all the reconstructed prompts.
DALL-V \cite{DALLV} employs large language-vision models for the task of source-free video domain adaptation. They distill the world prior and complementary source model information into a student network tailored for the target. This approach also freezes the vision encoder and only learns an adapter on top of the vision encoder.

Unlike most of the above methods, we do not freeze the vision encoder during the training. While most of the mentioned methods focus on adapting the text prompts, we focus on adapting the visual extractor while retaining the inherent knowledge through our strong-weak guidance. We also show our method is complementary to and benefits from prompt learning methods.

  \begin{table*}
  \caption{Accuracy results on Office-Home dataset. The best results are displayed in bold and the runner-up results are underlined. Methods using a ResNet-50 backbone are on top, and methods using a transformer-based backbone are on the bottom. CLIP indicates zero-shot results of CLIP, and methods based on it are listed below it. Our methods are on the bottom, DAPL marks the combination with the prompt learning method. V1 stands for version 1 and V2 for version 2 respectively.}
  \label{tab:officeHome}
  \centering
  \resizebox{\textwidth}{!}{
  \begin{tabular}{llllllllllllll}
    \toprule
    Method - RN50 & A$\shortrightarrow$C  & A$\shortrightarrow$P & A$\shortrightarrow$R & 
    C$\shortrightarrow$A & C$\shortrightarrow$P  & C$\shortrightarrow$R &
    P$\shortrightarrow$A & P$\shortrightarrow$C  & P$\shortrightarrow$R &
    R$\shortrightarrow$A & R$\shortrightarrow$C  & R$\shortrightarrow$P &
     Avg \\
    \midrule
    \midrule
	CDAN \cite{CDAN} & 50.7 & 70.6 & 76.0 & 57.6 & 70.0 & 70.0 & 57.4 & 50.9 & 77.3 & 70.9 & 56.7 & 81.6 & 65.8 \\
 	BIWAA-I \cite{BIWA} & 56.3 & {78.4} & {81.2} & 68.0 & 74.5 & 75.7 & 67.9 & 56.1 & 81.2 & 75.2 & 60.1 & 83.8 & 71.5 \\
 	Sentry \cite{SENTRY} & 61.8 & 77.4 & 80.1 & 66.3 & 71.6 & 74.7 & 66.8 & 63.0 & 80.9 & 74.0 & {66.3} & 84.1 & 72.2 \\
GSDE \cite{GSDE} & 57.8 & {80.2} & {81.9} & {71.3} & {78.9} & {80.5} & 67.4 & 57.2 & 84.0 & 76.1 & 62.5 & {85.7} & {73.6} \\
PCL \cite{PCL} & 60.8 & 79.8 & 81.6 & 70.1 & 78.9 & 78.9 & 69.9 & 60.7 & 83.3 & 77.1 & 66.4 & 85.9 & 74.5 \\
\midrule
DAPL \cite{DAPL} & 54.1 & 84.3 & 84.8 & 74.4 & 83.7 & 85.0 & 74.5 & 54.6 & 84.8 & 75.2 & 54.7 & 83.8 & 74.5 \\
AD-CLIP \cite{ADCLIP} & 55.4 & 85.2 & {85.6} & 76.1 & {85.8} & {86.2} & {76.7} & 56.1 & 85.4 & 76.8 & 56.1 & 85.5 & 75.9 \\
PADCLIP \cite{PADCLIP} & 57.5 & 84.0 & 83.8 & {77.8} & 85.5 & 84.7 & 76.3 & 59.2 & 85.4 & 78.1 & 60.2 & 86.7 & 76.6 \\
EUDA \cite{EMP} & 58.1 & {85.0} & 84.5 & 77.4 & 85.0 & 84.7 & 76.5 & 58.8 & {85.7} & 75.9 & 60.4 & 86.4 & 76.5 \\
\midrule
SWG V1 & 64.3 & 86.2 & 86.8 & 78.8 & 87.6 & 86.6 & 79.4 & 64.5 & 86.9 & 80.9 & 66.1 & 88.9 & 79.7 \\
SWG V2 & 66.6 & \underline{87.3} & 87.0 & 79.6 & 87.4 & \underline{86.7} & 78.8 & 66.5 & \underline{87.2} & 80.4 & 68.0 & 88.8 & 80.3 \\
SWG DAPL V1 & \underline{70.8} & \textbf{89.8} & \underline{88.9} & \underline{81.0} & \underline{89.9} & \textbf{89.0} & \underline{81.2} & \underline{70.8} & \textbf{89.3} & \underline{84.0} & \underline{74.0} & \textbf{92.2} & \underline{83.4} \\
SWG DAPL V2 & \textbf{73.8} & \textbf{89.8} & \textbf{89.3} & \textbf{81.7} & \textbf{90.0} & \textbf{89.0} & \textbf{81.6} & \textbf{74.4} & \textbf{89.3} & \textbf{84.7} & \textbf{76.4} & \underline{91.9} & \textbf{84.3} \\
\midrule
\midrule
    Method - ViT & A$\shortrightarrow$C  & A$\shortrightarrow$P & A$\shortrightarrow$R & 
    C$\shortrightarrow$A & C$\shortrightarrow$P  & C$\shortrightarrow$R &
    P$\shortrightarrow$A & P$\shortrightarrow$C  & P$\shortrightarrow$R &
    R$\shortrightarrow$A & R$\shortrightarrow$C  & R$\shortrightarrow$P &
     Avg \\
\midrule
CDAN \cite{CDAN} & 62.6 & 82.9 & 87.2 & 79.2 & 84.9 & 87.1 & 77.9 & 63.3 & 88.7 & 83.1 & 63.5 & 90.8 & 79.3 \\
SDAT \cite{SDAT} & 70.8 & 87.0 & 90.5 & 85.2 & 87.3 & 89.7 & 84.1 & 70.7 & 90.6 & 88.3 & 75.5 & 92.1 & 84.3 \\
SSRT \cite{SSRT} & 75.2 & 89.0 & 91.1 & 85.1 & 88.3 & 89.9 & 85.0 & 74.2 & 91.2 & 85.7 & 78.6 & 91.8 & 85.4 \\
PMTrans \cite{PMTrans} & 81.3 & 92.9 & 92.8 & 88.4 & 93.4 & 93.2 & {87.9} & 80.4 & \underline{93.0} & 89.0 & 80.9 & 94.8 & 89.0 \\
\midrule
DAPL \cite{DAPL} & 70.6 & 90.2 & 91.0 & 84.9 & 89.2 & 90.9 & 84.8 & 70.5 & 90.6 & 84.8 & 70.1 & 90.8 & 84.0 \\
AD-CLIP \cite{ADCLIP} & 70.9 & 92.5 & 92.1 & 85.4 & 92.4 & 92.5 & 86.7 & 74.3 & 93.0 & 86.9 & 72.6 & 93.8 & 86.1 \\
PADCLIP \cite{PADCLIP} & 76.4 & 90.6 & 90.8 & 86.7 & 92.3 & 92.0 & 86.0 & 74.5 & 91.5 & 86.9 & 79.1 & 93.1 & 86.7 \\
EUDA \cite{EMP} & 78.2 & 90.4 & 91.0 & 87.5 & 91.9 & 92.3 & 86.7 & 79.7 & 90.9 & 86.4 & 79.4 & 93.5 & 87.3 \\
\midrule
SWG V1 & 81.5 & 93.5 & 92.7 & 89.5 & 93.3 & 92.8 & 89.2 & 82.4 & 92.6 & 90.2 & 82.7 & 94.3 & 89.6 \\
SWG V2 & 81.6 & 93.6 & 92.7 & 89.6 & 93.3 & 93.1 & 89.1 & 83.1 & 92.9 & 90.4 & 83.1 & 94.1 & 89.7 \\
SWG DAPL V1 & \underline{85.8} & \underline{93.8} & \underline{93.5} & \underline{89.4} & \underline{94.0} & \textbf{93.9} & \underline{90.0} & \underline{86.8} & \textbf{94.2} & \underline{91.6} & \underline{87.4} & \underline{95.3} & \underline{91.3} \\
SWG DAPL V2 & \textbf{87.1} & \textbf{93.9} & \textbf{94.1} & \textbf{91.1} & \textbf{94.3} & \underline{93.8} & \textbf{91.1} & \textbf{89.0} & \textbf{94.2} & \textbf{92.3} & \textbf{89.4} & \textbf{95.8} & \textbf{92.2} \\
    \bottomrule
  \end{tabular}}
\end{table*}

\section{Methodology}

In unsupervised domain adaptation, the goal is to transfer knowledge from a source dataset $\mathcal D_s$ consisting of $n_s$ labeled images $\mathcal D_s = {(x_{i,s},y_{i,s})}_{i=1}^{n_s}$ to an unlabeled target dataset $ D_t$ containing $n_t$ unlabeled data $\mathcal D_t = {(x_{i,t})}_{i=1}^{n_t}$. In this work we focus on combining the inherent knowledge of vision-language models with the knowledge gained from the knowledge transfer of the source domain, meaning we further have access to the zero-shot predictions $y_{o}$ for each sample.

For the knowledge transfer from the source to the target domain, we employ the widely established adversarial domain adaptation method CDAN \cite{CDAN} as it was shown to be effective for both convolutional- and transformer-based network architectures \cite{pretrain}. 

To preserve the inherent knowledge of the vision-language model we employ a strong-weak guidance scheme. For the strong guidance, a hard pseudo-label is employed, but only for the most confident samples. We employ a source domain expansion inspired by GSDE \cite{GSDE} that copies the most confident target samples into the source domain with their predicted pseudo-label.

For the weak guidance, we employ a knowledge distillation loss \cite{kd} using the zero-shot predictions. However, since the zero-shot predictions exhibit a relatively wide spread of probabilities among the classes, we first accentuate the winning probabilities. The process flow of our algorithm can be seen in Fig. \ref{fig:flow}.

We use the three losses:
\begin{equation}
L=L_{CE} + L_{KD} + L_{AD}
\end{equation} 
where $L_{CE}$ is the classification loss for the source data, $L_{KD}$ is the knowledge distillation loss, and $L_{AD}$ is the adversarial adaptation loss.  $L_{KD}$ and $L_{AD}$ are calculated for both source and target data. We further employ a strongly augmented version of source and target data, which are handled in the same way as the weakly augmented versions. 

We also show that our method is complementary to prompt learning methods by combining it with Domain Adaptation via Prompt Learning (DAPL) \cite{DAPL}. 

\subsection{Strong guidance - Source Domain Expansion:}
For the strong guidance, we employ a source domain expansion strategy. The highest scoring zero-shot predictions are copied and appended to the source dataset with their respective pseudo-labels as pseudo-source data. We employ this source domain expansion strategy since it was shown to be more effective than simply adding a CE-loss for the respective target samples (as was shown in \cite{GSDE}). During training no distinction between source data and pseudo-source data is made.

The paper that inspired the strategy for our strong guidance employed the source domain expansion in an iterative procedure. The network was trained for several runs, each time with an increasing amount of pseudo-source data. 

In this work, we experiment with 2 different versions. The first version trains the network a single time. The second version employs an iterative training scheme training for 2 runs. In the first run, the source domain expansion is done regarding the zero-shot predictions, and in the second run it is done regarding a mixture of zero-shot predictions and predictions generated from the previous run:

\begin{equation} \label{eq:GSDE}
s(\hat y) = p(y_{n-1}) + \frac{1}{2} p(\tilde y_{o})
\end{equation}

where $s(\hat y)$ depicts the scores of the samples for being chosen as pseudo-source samples, $p(y_{n-1})$ depicts the predictions of the previous run, and $p(\tilde y_{o})$ depicts the zero-shot predictions.

While the first version trains for a single run, and therefore no additional computational time is required, the second version trains for two runs, which require double the training time.

\subsection{Weak guidance - Knowledge distillation loss}

For the weak guidance, we employ a knowledge distillation loss \cite{kd} using the zero-shot predictions. However, since the predictions are rather evenly spread among all classes, we accentuate the winning probabilities before employing them in the KD-loss (see Fig. \ref{fig:flow}). This is done through a temperature parameter $T$ in the softmax function. In a normal KD setting, the teacher's winning prediction score usually tends to approach a one-hot vector. The goal is to mimic this setting while maintaining the relative prediction probabilities. To accomplish that we estimate the temperature parameter $T$ so that the mean winning probability over all source and target data equals to $\tau$. We employ a value of $\tau = 0.9$ in our experiments. In the estimation, we equally factor the source and target domain to further maintain the relative prediction confidence between the two domains.

\begin{equation}
\frac{1}{2n_s} \sum_{n_s} \max(\sigma(y_{o,i,s} / T)) + \frac{1}{2n_t} \sum_{n_t} \max(\sigma(y_{o,i,t} / T)) = \tau
\end{equation} 

$y_{o,i,s}$ depicts logits of the $i$-th sample of the $s$ource data the $\sigma$ depicts the softmax-function. 
We then employ the adjusted zero-shot predictions for the knowledge distillation loss:

\begin{equation}
L_{KD} = D_{KL}(\tilde y_{o} || y)
\end{equation} 

where $D_{KL}$ is the Kullback-Leibler divergence loss, and $y$ is the output of the classifier of our network.

  
\subsection{Adversarial loss}
We employ conditional adversarial domain adaptation loss CDAN:
\begin{equation}
L_{AD} = L_{BCE}(G_d((f_i \otimes p_i), d_i))
\end{equation}
where $G_d$ is the domain classification network. $f_i$ are the features of sample $i$, $p_i$ the class probabilities and $d_i$ the domain label. CDAN employs a gradient reversal layer to inverse the training objective. This means that while the domain classifier is trained to distinguish the domain (source or target) of the respective sample, the objective of the feature extractor is to extract features that are indistinguishable for the domain classifier, thereby generating domain invariant features.

  \begin{table*}
  \caption{Accuracy results on VisDA dataset. The best results are displayed in bold and the runner-up results are underlined. Methods using a ResNet-101 backbone are on top, and methods using a transformer-based backbone are on the bottom. CLIP indicates zero-shot results of CLIP, and methods based on it are listed below it. Our methods are on the bottom, DAPL marks the combination with the prompt learning method. V1 stands for version 1 and V2 for version 2 respectively.}
  \label{tab:visda}
  \centering
  \resizebox{\textwidth}{!}{
  \begin{tabular}{llllllllllllll}
    \toprule
    Method - RN101 & plane & bcycl & bus & car & horse & knife & mcycl & person & plant & sktbrd & train & truck &  Avg \\
    \midrule
CDAN \cite{CDAN} & 85.2 & 66.9 & 83.0 & 50.8 & 84.2 & 74.9 & 88.1 & 74.5 & 83.4 & 76.0 & 81.9 & 38.0 & 73.9 \\
STAR \cite{STAR} & 95.0 & 84.0 & 84.6 & 73.0 & 91.6 & 91.8 & 85.9 & 78.4 & 94.4 & 84.7 & 87.0 & 42.2 & 82.7 \\
CAN \cite{CAN} & 97.0 & 87.2 & 82.5 & 74.3 & 97.8 & \underline{96.2} & 90.8 & 80.7 & \textbf{96.6} & \underline{96.3} & 87.5 & 59.9 & 87.2 \\
CoVi \cite{CoVi} & 96.8 & 85.6 & 88.9 & \textbf{88.6} & 97.8 & 93.4 & 91.9 & \textbf{87.6} & \underline{96.0} & 93.8 & 93.6 & 48.1 & 88.5 \\
\midrule
DAPL \cite{DAPL} & 97.8 & 83.1 & 88.8 & 77.9 & 97.4 & 91.5 & 94.2 & 79.7 & 88.6 & 89.3 & 92.5 & 62.0 & 86.9 \\
AD-CLIP \cite{ADCLIP} & 98.1 & 83.6 & 91.2 & 76.6 & 98.1 & 93.4 & {96.0} & 81.4 & 86.4 & 91.5 & 92.1 & 64.2 & 87.7 \\
PADCLIP \cite{PADCLIP} & 96.7 & \underline{88.8} & 87.0 & 82.8 & 97.1 & 93.0 & 91.3 & 83.0 & 95.5 & 91.8 & 91.5 & 63.0 & 88.5 \\
EUDA \cite{EMP} & 97.2 & \textbf{89.3} & 87.6 & \underline{83.1} & \underline{98.4} & 95.4 & 92.2 & 82.5 & 94.9 & 93.2 & 91.3 & 64.7 & 89.2\\
\midrule
SWG V1  & \textbf{99.0} & 85.2 & 91.3 & 72.9 & \textbf{98.8} & \textbf{96.4} & \underline{96.2} & 83.0 & 91.0 & 96.2 & 94.0 & \textbf{75.0} & 89.9 \\
SWG V2  & \underline{98.9} & 87.2 & 91.6 & 73.8 & \textbf{98.8} & 96.1 & \textbf{96.3} & \underline{83.2} & 91.8 & 95.7 & \underline{94.2} & \underline{74.5} & \textbf{90.2} \\
SWG DAPL V1 & \underline{98.9} & 87.3 & \textbf{92.8} & 81.1 & 98.0 & 95.6 & 96.0 & 82.8 & 92.4 & \underline{96.3} & 93.1 & 66.4 & \underline{90.1} \\
SWG DAPL V2 & \underline{98.9} & 87.5 & \underline{92.3} & 82.5 & 98.3 & 94.2 & 95.7 & 80.7 & 91.8 & \textbf{96.8} & \textbf{94.5} & 66.0 & 89.9 \\
    \midrule
    \midrule
    Method - ViT & plane & bcycl & bus & car & horse & knife & mcycl & person & plant & sktbrd & train & truck &  Avg \\
    \midrule
PMTrans \cite{PMTrans} & 99.4 & 88.3 & 88.1 & {78.9} & 98.8 & {98.3} & 95.8 & 70.3 & 94.6 & 98.3 & 96.3 & 48.5 & 88.0 \\
SSRT \cite{SSRT} & 98.9 & 87.6 & 89.1 & 84.8 & 98.3 & \textbf{98.7} & {96.3} & 81.1 & {94.8} & 97.9 & 94.5 & 43.1 & 88.8 \\
AdaCon \cite{AdaCon} & 99.5 & 94.2 & 91.2 & 83.7 & 98.9 & 97.7 & {96.8} & 71.5 & \underline{96.0} & 98.7 & \textbf{97.9} & 45.0 & 89.2 \\
DePT-D \cite{DePT} & 99.4 & 93.8 & \underline{94.4} & \textbf{87.5} & \underline{99.4} & {98.0} & {96.7} & 74.3 & \textbf{98.4} & 98.5 & {96.6} & 51.0 & 90.7 \\
\midrule
DAPL \cite{DAPL} & 99.2 & 92.5 & 93.3 & 75.4 & 98.6 & 92.8 & 95.2 & 82.5 & 89.3 & 96.5 & 95.1 & 63.5 & 89.5 \\
AD-CLIP \cite{ADCLIP} & 99.6 & 92.8 & 94.0 & 78.6 & 98.8 & 95.4 & {96.8} & 83.9 & 91.5 & 95.8 & 95.5 & 65.7 & 90.7 \\
PADCLIP \cite{PADCLIP} & 98.1 & 93.8 & 87.1 & \underline{85.5} & 98.0 & 96.0 & 94.4 & 86.0 & 94.9 & 93.3 & 93.5 & 70.2 & 90.9 \\
EUDA \cite{EMP} & 98.4 & 94.3 & 89.0 & 85.4 & 98.5 & {98.3} & 96.1 & \underline{86.3} & 95.1 & 95.2 & 92.5 & \underline{70.9} & {91.7} \\
\midrule
SWG V1 & \textbf{99.8} & 93.8 & 93.7 & 73.7 & \textbf{99.6} & 98.3 & \underline{97.3} & 82.1 & 91.0 & \underline{99.1} & \underline{96.8} & 70.0 & 91.3 \\
SWG V2 & \underline{99.7} & 94.0 & 93.6 & 73.7 & \textbf{99.6} & \underline{98.4} & \textbf{97.5} & 81.4 & 92.2 & \textbf{99.3} & 96.7 & \textbf{71.4} & 91.5 \\
SWG DAPL V1 & 99.6 & \underline{94.9} & \textbf{94.6} & 78.3 & \underline{99.1} & \underline{98.4} & 96.7 & 85.9 & 92.7 & 99.1 & 95.8 & 69.4 & \underline{92.0} \\
SWG DAPL V2 & 99.5 & \textbf{95.5} & \textbf{94.6} & 79.1 & 99.1 & 97.8 & 96.9 & \textbf{86.5} & 92.8 & \underline{99.1} & 95.9 & 70.7 & \textbf{92.3} \\

    \bottomrule
  \end{tabular}}
\end{table*}

  \begin{table*}
  \caption{Accuracy results on DomainNet dataset.}
  \label{tab:domainNet}
  \centering
  \resizebox{\textwidth}{!}{
  \begin{tabular}{c|ccccccc||c|ccccccc||c|ccccccc}
    \toprule
MCD & clp & inf & pnt & qdr & rel & skt & Avg & SCDA & clp & inf & pnt & qdr & rel & skt & Avg&CDTrans & clp & inf & pnt & qdr & rel & skt & Avg \\
\midrule
clp & - &15.4&25.5&3.3&44.6&31.2&24.0& clp & - &18.6&39.3&5.1&55.0&44.1&32.4&clp & - &29.4&57.2&26.0&72.6&58.1&48.7\\
inf &24.1& - &24.0&1.6&35.2&19.7&20.9& inf &29.6& - &34.0&1.4&46.3&25.4&27.3&inf &57.0& - &54.4&12.8&69.5&48.4&48.4\\
pnt &31.1&14.8& - &1.7&48.1&22.8&23.7& pnt &44.1&19.0& - &2.6&56.2&42.0&32.8&pnt &62.9&27.4& - &15.8&72.1&53.9&46.4\\
qdr &8.5&2.1&4.6& - &7.9&7.1&6.0& qdr &30.0&4.9&15.0& - &25.4&19.8&19.0&qdr &44.6&8.9&29.0& - &42.6&28.5&30.7\\
rel &39.4&17.8&41.2&1.5& - &25.2&25.0& rel &54.0&22.5&51.9&2.3& - &42.5&34.6&rel &66.2&31.0&61.5&16.2& - &52.9&45.6\\
skt &37.3&12.6&27.2&4.1&34.5& - &23.1& skt &55.6&18.5&44.7&6.4&53.2& - &35.7&skt &69.0&29.6&59.0&27.2&72.5& - &51.5\\
Avg &28.1&12.5&24.5&2.4&34.1&21.2&\cellcolor[gray]{0.8}20.5& Avg &42.6&16.7&37.0&3.6&47.2&34.8&\cellcolor[gray]{0.8}30.3&Avg &59.9&25.3&52.2&19.6&65.9&48.4&\cellcolor[gray]{0.8}45.2\\
\midrule
\midrule
 SSRT & clp & inf & pnt & qdr & rel & skt & Avg & PMTrans & clp & inf & pnt & qdr & rel & skt & Avg&EUDA & clp & inf & pnt & qdr & rel & skt & Avg \\
 \midrule
 clp & - &33.8&60.2&19.4&75.8&59.8&49.8& clp & - &34.2&62.7&32.5&79.3&63.7&54.5&clp & - &54.8&69.9&35.3&85.1&67.4&62.5\\
 inf &55.5& - &54.0&9.0&68.2&44.7&46.3& inf &67.4& - &61.1&22.2&78.0&57.6&57.3&inf &70.2& - &68.5&16.6&82.2&65.9&60.7\\
 pnt &61.7&28.5& - &8.4&71.4&55.2&45.0& pnt &69.7&33.5& - &23.9&79.8&61.2&53.6&pnt &72.4&54.6& - &29.5&83.0&66.4&61.2\\
 qdr &42.5&8.8&24.2& - &37.6&33.6&29.3& qdr &54.6&17.4&38.9& - &49.5&41.0&40.3&qdr &73.1&50.8&64.3& - &81.2&62.3&66.3\\
 rel &69.9&37.1&66.0&10.1& - &58.9&48.4& rel &74.1&35.3&70.0&25.4& - &61.1&53.2&rel &75.5&56.1&74.6&30.2& - &65.6&60.4\\
 skt &70.6&32.8&62.2&21.7&73.2& - &52.1& skt &73.8&33.0&62.6&30.9&77.5& - &55.6&skt &74.9&56.2&70.2&32.3&80.3& - &62.8\\
 Avg &60.0&28.2&53.3&13.7&65.3&50.4&\cellcolor[gray]{0.8}45.2& Avg &67.9&30.7&59.1&27.0&72.8&56.9&\cellcolor[gray]{0.8}52.4&Avg &73.2&54.5&69.5&28.8&82.4&65.5&\cellcolor[gray]{0.8}62.3\\
 \midrule
 \midrule
 PADCLIP & clp & inf & pnt & qdr & rel & skt & Avg & SWG V1 & clp & inf & pnt & qdr & rel & skt & Avg& SWG V2 & clp & inf & pnt & qdr & rel & skt & Avg\\
 \midrule
 clp & - &55.1&71.1&36.8&84.2&68.1&63.1& clp & - &58.4&73.1&25.7&85.0&71.9&62.8& clp & - &58.2&74.0&30.4&85.4&72.7&64.2\\
 inf &73.6& - &70.6&18.0&83.5&66.6&62.5& inf &78.7& - &73.1&24.7&84.8&70.8&66.4& inf &79.7& - &73.7&28.5&85.3&71.2&67.7\\
 pnt &75.4&54.3& - &32.0&83.5&67.2&62.5& pnt &79.6&58.3& - &26.0&84.6&71.3&63.9& pnt &80.3&58.4& - &27.8&84.8&71.9&64.6\\
 qdr &74.6&53.6&70.0& - &83.1&66.1&69.5& qdr &76.5&56.2&70.6& - &84.0&69.2&71.3& qdr &76.3&53.5&71.4& - &83.8&69.6&70.9\\
 rel &76.4&54.9&72.7&31.7& - &67.5&60.6& rel &80.4&58.5&73.9&24.4& - &71.4&61.7& rel &81.2&58.8&74.4&29.6& - &72.0&63.2\\
 skt &76.3&54.9&71.7&34.9&83.6& - &64.3& skt &80.6&58.1&73.6&25.1&84.8& - &64.5& skt &81.6&58.7&74.8&30.5&85.3& - &66.2\\
 Avg &75.3&54.6&71.2&30.7&83.6&67.1&\cellcolor[gray]{0.8}63.7& Avg &79.2&57.9&72.9&25.2&84.6&70.9&\cellcolor[gray]{0.8}65.1& Avg &79.8&57.5&73.7&29.4&84.9&71.5&\cellcolor[gray]{0.8}66.1\\
\bottomrule
  \end{tabular}}
\end{table*}

\subsection{Further improvements:}

\subsubsection{Batch norm layer adjustment}
The ResNet backbone employs batch norm layers. However, the distribution of the pretraining data is vastly different than the data used for the domain adaptation. 
To overcome this problem we employ domain-specific BN layers as in \cite{DBN} and adjust the BN layers to the statistics of the data of their respective domain similar to \cite{BNAD}. We estimate the average and mean for the source and target data for each batch norm layer and adjust the learnable parameters $\beta$ and $\gamma$ so that the running mean and var equals that of our training data (one adjusted to source domain and one to the target domain).

\begin{equation}
\tilde{\beta} = \beta - (\mu_p - \mu_c) * \frac{\gamma}{\sqrt \sigma_p})
\end{equation}

\begin{equation}
\tilde{\gamma} = (\gamma * \frac{\sqrt{\sigma_c}}{\sqrt{\sigma_p}})
\end{equation}

where $\mu_p$ and $\sigma_p$ represent the running mean and variance of the pre-trained model, and $\mu_c$ and $\sigma_c$ represent the estimated mean and variance of the training data. We want to highlight that this is only done for the ResNet backbone since the ViT backbone employs Layer Normalization instead of BN. 

\subsubsection{Zero-shot predictions}
To increase the quality of zero-shot predictions, we try to keep the original ratios of the images. The smaller side of an image gets resized to 224 pixels and the larger side gets resized by the same factor but rounded to the closes multiple of 16 - according to the patchsize of the ViT-backbone (32 for ResNet-backbone due to the attention pooling). We interpolate the positional embedding in order for the network to process the different image sizes.

\subsubsection{DAPL}
Domain Adaptation via Prompt Learning (DAPL) \cite{DAPL} focuses on adapting the prompts of the text encoder. It freezes both text and visual encoder during the training. In contrast to this, our method focuses on adapting the visual encoder. We combine these two methods in a sequential manner. In a first step, we run the DAPL algorithm and use the adapted text prompts to estimate the probabilities for the source and target data, which are then employed instead of the zero-shot predictions.

\section{Experiments}
We evaluate our proposed method on three different domain adaptation benchmarks, Office-Home, VisDA, and DomainNet. We show that we can improve the baselines significantly.

 \begin{figure}[t]
   \centering
                \includegraphics[width=0.4\textwidth ]{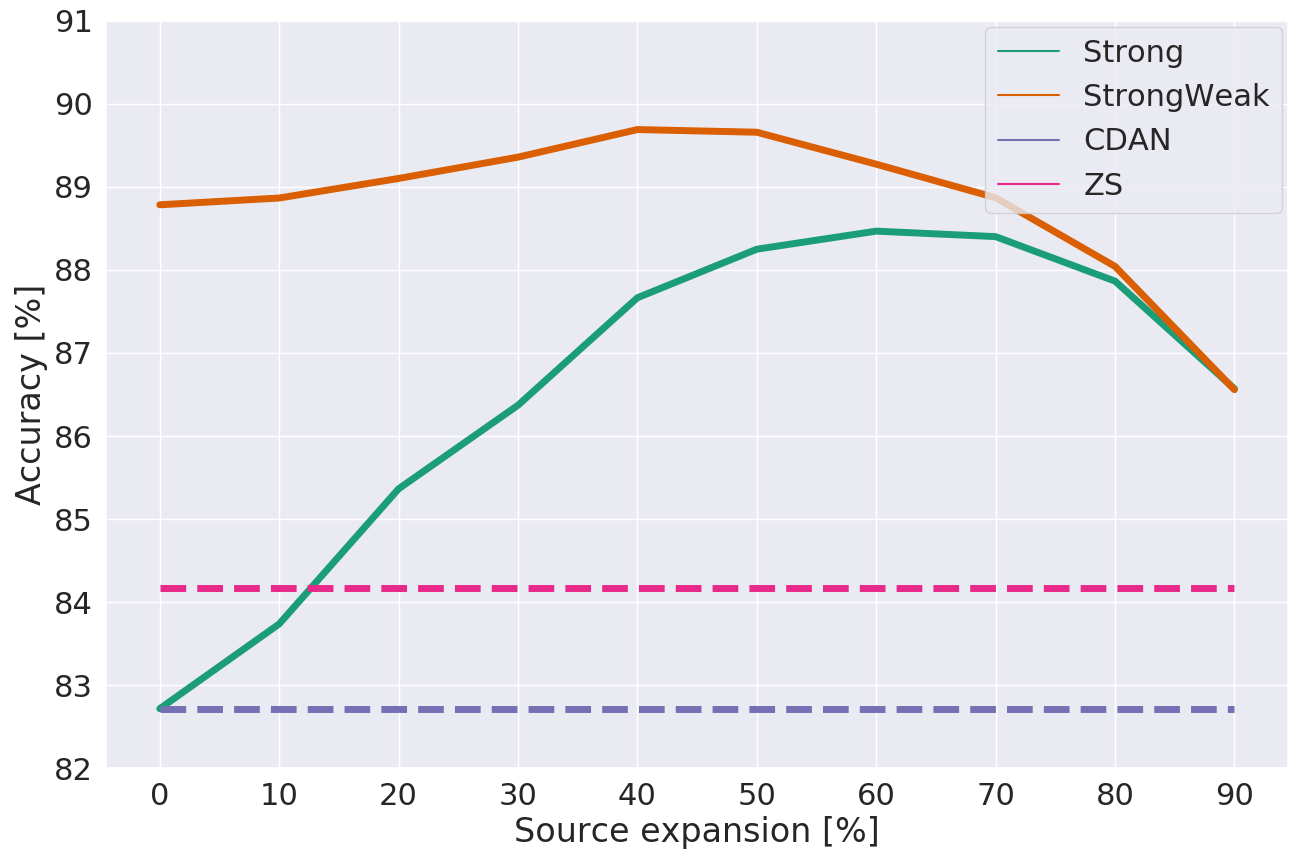}
                \vspace{-.3cm}
                \caption{Accuracy for adaptation of OfficeHome dataset for different source domain expansion percentages. The green line employs only the strong guidance, while the orange line employs both guidance. Additionally, the baselines of only using the adversarial domain adaptation (CDAN) and the CLIP zero-shot accuracy (ZS) are plotted.}
                \label{fig:sw}      
                \vspace{-.3cm} 
  \end{figure}
  
\subsection{Experiment settings}

\textbf{Office-Home} \cite{OH} contains 15,500 images of 65 categories. The domains are Art (A), Clipart (C), Product (P), and Real-World (R). We evaluate all twelve possible adaptation tasks. 

\textbf{VisDA} \cite{visda} is a challenging dataset for for synthetic-to-real domain adaptation. The synthetic source domain consists of 152397 images and the real-world target domain consists of 55388 images over 12 different classes.

\textbf{DomainNet} \cite{DN} is a large-scale dataset with about 600,000 images from 6 different domains and 345 different classes. We evaluate all 30 possible adaptation tasks. 

\textbf{Implementation details:} 
We built up our implementation on the CDAN implementation of \cite{CDAN}. We do all experiments with the CLIP-trained versions of the ResNet backbone and transformer backbone. For the ResNet backbone, we follow most other publications and employ ResNet-50 for OfficeHome and ResNet-101 for VisDA. For the transformer backbone, we employ ViT-B/16. We employ the seven ImageNet templates subset of the CLIP Git Hub page as text prompts for estimating the zero-shot predictions. 
For OfficeHome we train each run for $100$ episodes, for VisDA and DomainNet we train for $50$ episodes. For OfficeHome and DomainNet, we employ AdamW optimizer with a learning rate of $5e^{-6}$ for the backbone and $5e^{-5}$ for the newly initialized layers. For VisDA we employ a lower learning rate of $1e^{-6}$, and $1e^{-5}$ respectively. 
We employ a batch size of $64$. We load each image once with weak and once with strong data augmentation, but don't discriminate between the augmentations in the processing pipeline. 
We run two different versions of our method. The first version (V1) trains with a source expansion set to $50\%$. The second version (V2) trains at first with a source expansion set to $33.3\%$ in a first run, and $66.6\%$ in a second run, where the first run employs the zero-shot predictions and the second a mixture between zero-shot and previous run predictions for the source expansion. Due to the two training runs, the second version requires about double the training time. 
We employ the official implementation of DAPL for the respective experiments.

 \begin{figure}[t]
   \centering
                \includegraphics[width=0.4\textwidth ]{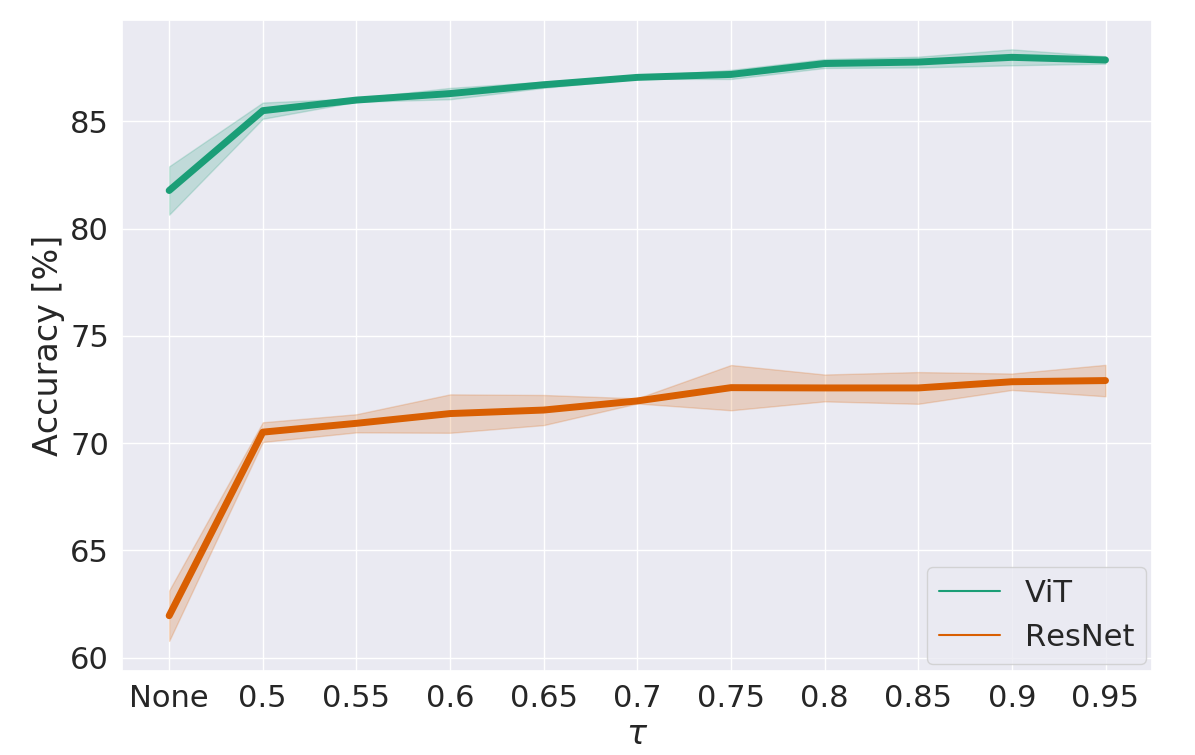}
                \vspace{-.3cm}
                \caption{Accuracy for adaptation of OfficeHome C$\rightarrow$A for different values for $\tau$. None represents directly using the zero-shot predictions without adjusting the output probabilities. It can be seen that adjusting the probability distribution is fundamental for the knowledge distillation to work properly.}
                \label{fig:tau}       
                \vspace{-.3cm}
  \end{figure}

\subsection{Results}
\textbf{Results for Office-Home:} The results for the Office-Home dataset are shown in Tab. \ref{tab:officeHome}. All versions of our method vastly outperform earlier work. The addition of DAPL improves the results by around $3-4\%$pts in the case of the ResNet backbone and around $2\%$pts for the ViT backbone. The second version using an iterative training scheme with 2 runs increases the accuracy in all cases, this comes however at the cost of roughly double the training time.

\textbf{Results for VisDA-2017:} The results for the VisDA-2017 dataset are shown in Tab. \ref{tab:visda}. Our method just by itself already outperforms most of the earlier works. Our method slightly performs worse than EUDA only for the case of the ViT-backbone and without using DAPL. However, once we add DAPL, we significantly outperform the next best-performing method for the ResNet and ViT backbone. 

\textbf{Results for DomainNet:} The results for the DomainNet dataset are shown in Tab. \ref{tab:domainNet}. The CLIP-based adaptation methods clearly outperform conventional methods by upwards of $10\%$pts. Our method significantly outperforms current state-of-the-art methods, increasing the average accuracy by $1.4\%$pts or $2.4\%$pts respectively over the next best performing method. Version 2, the iterative version of our method increases the accuracy by $1\%$pts over version 1. This however comes again at the price of the increase in training time.

\section{Ablation studies}

\begin{table}
  \caption{Effect of batch-norm layer adjustment. BN layer non adjusted \textbf{(C)}, BN layer adjusted for source data \textbf{(C-SAd)}, adjusted for target data \textbf{(C-TAd)}, domain specific BN non adjusted \textbf{(D)}, and domain specific layer adjusted to their specific domain \textbf{D-STAd}. \vspace*{-.25cm}}
  \label{tab:BN}
  \centering
  \begin{tabular}{l|l|l|l|l|l}
    Method & S & S-SAd & S-TAd & D & D-STAd \\
    \midrule
    OH & 78.5 & 78.8 & 78.9 & 78.5 & 79.7 \\
  \end{tabular}
  \vspace*{-.25cm}  
\end{table}

\textbf{Combination of strong and weak guidance}
The effect of combining the strong and weak guidance can be seen in Fig. \ref{fig:sw}. It should be noted in the case of $0\%$ source expansion, the strong guidance is equivalent to CDAN, and the strong-weak guidance is equivalent to only weak guidance. The guidance vastly improves over the baselines, and the combination of both even further improves the results. According to the results we chose a source expansion of $50\%$ in the single run experiments of our evaluations.

\textbf{Parameter for shifting the zero-shot predictions $\tau$:}
The effect of the parameter $\tau$ is shown in Fig. \ref{fig:tau}. None means that no adjustment was applied to the zero-shot predictions. It can be seen that applying the ZS probably adjustment is essential for the method. Employing a target mean winning probability of $\tau = 0.9$ increases the accuracy by $10.9$pp for ResNet and $6.2$pp for ViT backbone. For the ViT backbone, the accuracy increases constantly until a value of $0.9$, for which the accuracy peaks. For the ResNet backbone, it seems that even higher values of $\tau$ yield better results. In all our experiments, we employ a value for $\tau$ of $0.9$.

\textbf{Batchnorm layer adjustment:} 
In Tab. \ref{tab:BN} it can be seen that adjusting the BN layer to either source or target gains a small increase in performance. Changing the BN to domain specific BN does not improve the performance just by itself, the performance increases just in combination with adjusting each BN layer to its domain. The performance gain of $1.2\%$pts is quite considerable. But even without adjusting the BN layer, we still outperform all other methods.

\textbf{Combining the weak-strong guidance with other DA methods:} 
In this part, we evaluated how well the weak-strong guidance fares other DA methods. We did not employ BN adjustment for this experiment. The results are shown in Tab. \ref{tab:meth}. While MDD outperforms CDAN for the ResNet backbone, it underperforms for the transformer backbone. For ViT, choosing the transformer-based method SSRT performs best. CDAN performs good for both backbones.


\textbf{Influence of quality of ZS predictions:} 
In this part, we evaluated the influence of the quality of ZS predictions on the adaptation. We generated different ZS predictions with a combination of different text templates (\textbf{s}ingle vs \textbf{m}ultiple) and image size (\textbf{r}ectangular vs ratio \textbf{p}reserving). Single template is "a + {cls}" while multiple is the ImageNetSmall list from CLIP. Rectangular image size resizes the image to 224x224, while ratio preserving resizes the smaller side to 224 and the larger side to its ratio preserving size (it uses adaptive input size). It can be seen (Tab. \ref{tab:zs}) that the quality of the zero-shot prediction has a big influence. However, while the difference between best and worst ZS prediction is $5\%$pts (ResNet); $3.8\%$pts (ViT), it's only $2.4\%$pts; $1\%$pts after the adaption. 

\begin{table}
  \caption{Combination of weak-strong guidance with other DA methods. The results shown are for the OfficeHome dataset. Besides CDAN, we employed MDD, SENTRY, and SSRT (transformer-based method).  \vspace*{-.25cm}  }
  \label{tab:meth}
  \centering
  \begin{tabular}{l|l|l|l|l}
    Method & CDAN & MDD & SENTRY & SSRT\\
    \midrule
    ResNet & \underline{78.5} & \textbf{80.0} & 78.2 & -  \\
    ViT & \underline{89.7} & 87.9 & 85.7 & \textbf{90.1}  \\
  \end{tabular}
  \vspace*{-.25cm}  
\end{table}

\begin{table}
  \caption{Effect of ZS prediction quality on adaptation. The ZS predictions are generated with different combinations of text templates (\textbf{S}ingle vs \textbf{M}ultiple) and image size (\textbf{R}ectangular vs ratio \textbf{P}reserving). \vspace*{-.25cm}}
  \label{tab:zs}
  \centering
  \begin{tabular}{l|l|l|l|l|l}
     &  & S-R & M-R & S-P & M-P\\
    \midrule
    ResNet & ZS & 70.3 & 73.5 & 73.0 & 75.3  \\
     & Results & 77.3 & 78.4 & 78.8 & 79.7 \\
     \midrule
    ViT & ZS & 80.4 & 83.2 & 83.2 & 84.2  \\
     & Results & 88.7 & 89.3 & 89.1 & 89.7 \\
  \end{tabular}
  \vspace*{-.25cm}  
\end{table}

\section{Conclusion}
In this work, we presented a novel way to combine the knowledge of vision-language models with knowledge gained from a source dataset through a strong-weak guidance scheme. 
For the strong guidance, we expand the source dataset with the most confident samples of the target dataset. Additionally, we employ a knowledge distillation loss as weak guidance.
The strong guidance uses hard labels but is only applied to the most confident predictions from the target dataset. Conversely, the weak guidance is employed to the whole dataset but uses soft labels. 
For the domain adaptation loss, we employed CDAN as it is effective for both CNN and ViT-based backbones, but also show that other methods can be used. Furthermore, we showed that our method is complementary to prompt learning methods and a combination can further improve the accuracy. 

\section*{Acknowledgment}
This work was partially supported by JST Moonshot R\&D Grant Number JPMJPS2011, CREST Grant Number JPMJCR2015 and Basic Research Grant (Super AI) of Institute for AI and Beyond of the University of Tokyo.

{\small
\bibliographystyle{ieee_fullname}
\bibliography{egbib}
}

\end{document}